\documentclass[conference]{IEEEtran}

\usepackage{cite}
\usepackage{amsmath,amssymb,amsfonts}
\usepackage{algorithmic}
\usepackage{graphicx}
\usepackage{textcomp}
\usepackage{xcolor}

\usepackage{subfigure}
\long\def\invis#1{}
\usepackage{hyphenat}
\usepackage{wrapfig,url}
\usepackage{siunitx}

\renewcommand{\figurename}{Figure}
\newcommand\etal{\textit{et al.\ }}

\newcommand\eg{\textit{e}.\textit{g}., }

\begin{document}
\date{}

\title{\LARGE \bf An Autonomous Surface Vehicle for Long Term Operations}
\author{Jason Moulton$^1$, Nare Karapetyan$^1$, Sharon Bukhsbaum$^1$, Chris McKinney$^1$, \\Sharaf Malebary$^2$, George Sophocleous$^1$, Alberto Quattrini Li$^3$, and Ioannis Rekleitis$^1$\\
$^1$Computer Science and Engineering Department, University of South Carolina\\
$^2$Information Technology Department, King Abdulaziz University\\
$^3$Computer Science Department, Dartmouth College
}
\maketitle

\normalsize

\begin{abstract}
Environmental monitoring of marine environments presents several challenges: the harshness of the environment, the often remote location, and most importantly, the vast area it covers. Manual operations are time consuming, often dangerous, and labor intensive. Operations from oceanographic vessels are costly and limited to open seas and generally deeper bodies of water. In addition, with lake, river, and ocean shoreline being a finite resource, waterfront property presents an ever increasing\hyp valued commodity, requiring exploration and continued monitoring of remote waterways. In order to efficiently explore and monitor currently known marine environments as well as reach and explore remote areas of interest, we present a design of an autonomous surface vehicle (ASV) with the power to cover large areas, the payload capacity to carry sufficient power and sensor equipment, and enough fuel to remain on task for extended periods. An analysis of the design and a discussion on lessons learned during deployments is presented in this paper.

\end{abstract}

\section{Introduction}
The University of South Carolina's Autonomous Field Robotics Lab (AFRL) Jetyak is an ASV modeled after the Woods Hole Oceanographic Institution (WHOI) Jetyak\cite{whoiMokai2014}. This work focuses on improving modularity and performance throughout the design and build phases in order to expand capabilities for operating in different environments. Furthermore, 
\invis{\textcolor{red}{N: Our design and implementation diverges from their ASV} \textcolor{blue} {N:} } the proposed design and implementation aims to expand deployment capabilities to include highly dynamic environments typically occurring in remote, uninhabited areas. \invis{Throughout our development, we define these dynamics as wind, water currents, and depths.} Along with our desire to maintain an on\hyp board manual operation mode, this expansion is guided in increasing the diversity of the operating modes and payloads, by setting the modularity and control as core implementation requirements of the platform.  

The ASV described in this paper is based on the Mokai Es\hyp Kape~\cite{mokai} boat. It is controlled using a Pixhawk PX4 micro\hyp controller, and is capable of communicating using \SI{900}{\MHz} radio modems, \SI{2.4}{\GHz} remote control radios, and a \SI{2.4}{\GHz} WiFi connection. The communication capabilities enable connectivity with: a remote control transmitter, a remote computer termed Ground Control Station (GCS), and other ASVs using an ad\hyp hoc network. The proposed design enables the following operating modes:
\begin{figure}[t]
		\centering      
		\includegraphics[width=1\columnwidth]{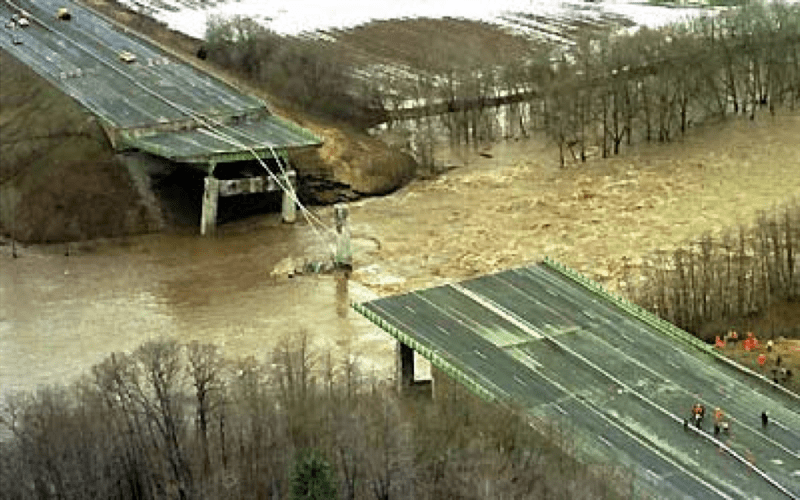}
		\caption{Example of a highly dynamic environment ideal for ASV deployment for the task of infrastructure inspection.}
		\label{fig:bridge}
	\end{figure}

\begin{itemize}
\item {\bf Manual operation on\hyp board:} A human uses the manual control of the Mokai Es\hyp Kape~\cite{mokai} to drive the vessel. This mode is valuable for a scientist to manually drive to a location and collect data, as well to test the dynamics of the vessel in challenging conditions. Furthermore, this mode can be utilized for learning via demonstration of autonomous control. 
\item {\bf Manual operation off\hyp board:} A human uses a \SI{2.4}{\GHz} remote control radio to operate the vehicle. Such mode can be employed to send the vessel to collect data in a challenging situation, especially when such operation raises safety concerns for a human operator.
\item {\bf Autonomous way\hyp point operation off\hyp board:} The boat is sent GPS way\hyp points via a remote computer. That way a single GCS can control multiple vehicles and coordinate with respect to collision avoidance. 
\item {\bf Autonomous way\hyp point operation on\hyp board:} A computer on\hyp board sends GPS way\hyp points to the vehicles micro\hyp controller (Pixhawk PX4). Decisions are made locally, and the vehicle can operate even if the communication with the GCS is intermittent.
\item {\bf Autonomous velocity control on/off\hyp board:} A computer uses a control algorithm (PID, adaptive, or model based) to change the steering angle and the forward velocity of the vessel based on sensory input. Such capability is critical for operating in adversarial conditions, such as high currents and strong winds.
\end{itemize}

\begin{figure*}[t]
  \includegraphics[width=1\textwidth,height=9cm]{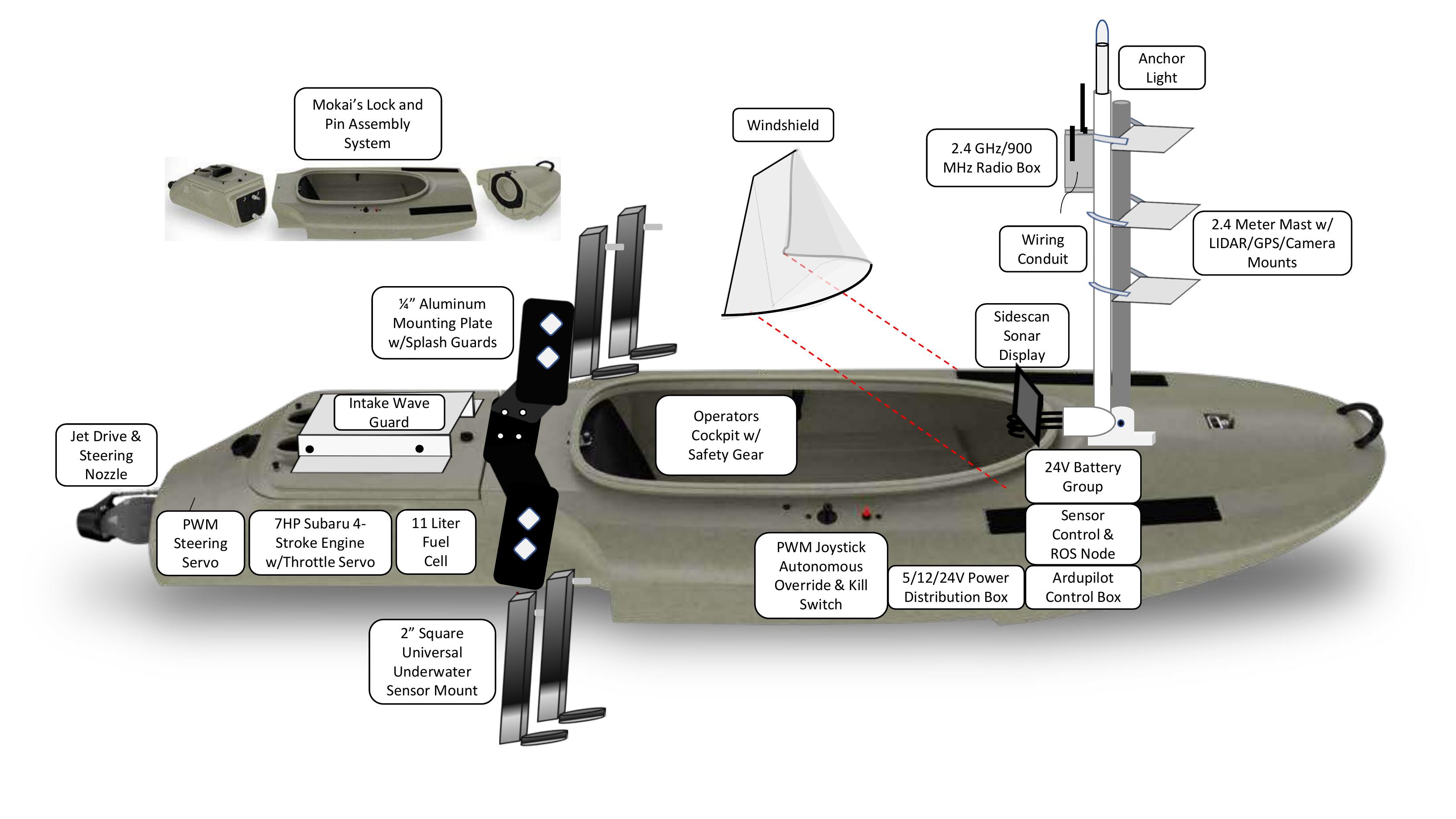}
  \caption{Stock Mokai Es\hyp Kape with additional modular components added by UofSC's AFRL to accommodate a multitude of underwater and above the surface sensors. Splash guards added in 3 locations to protect engine and on\hyp board electronics. Navigation lights and safety equipment are included on each JetYak for Coast Guard compliance. }
  \label{fig:usc_jetyak}
\end{figure*}
    
The main contributions of this paper lie, first, on expanding the modularity and flexibility of existing ASV platforms; providing a modular design of an ASV with publicly available documentation and software~\cite{doc}; discussing lessons learned during the construction of the vehicle and various deployments; and demonstration of our implementation's expanded capabilities through providing preliminary data collected in stable and highly dynamic environments.

The paper is structured as follows. Section \ref{sec:background} discusses related work, current design goals, and where the design goals diverge between previous approaches and the proposed design. Section \ref{sec:usc_build} details the construction process of the base platform.  Section \ref{sec:lessons} presents additional lessons learned not covered in the build section as well as valuable lessons learned through over 50 deployments of up to four Jetyaks simultaneously. Section \ref{sec:results} illustrates the current navigation and data collection success of this platform. Section \ref{sec:future_work} describes the ongoing work that AFRL is planning for increasing  the platform's capabilities, and finally we will conclude with Section \ref{sec:conclusion}.

\section{Background} \label{sec:background}
\subsection{Related Work \& Current State of The Art}
\invis{\textcolor{red}{N: [Whoi work isn't discussed at this point] In addition to the WHOI Jetyak, referenced and credited throughout this paper},}

The most relevant work to the proposed design is the WHOI Jetyak by Kimball \etal \cite{whoiMokai2014}, on which we have based our design. Their Jetyak is capable of autonomous operations carrying different payloads for extended periods of time; however, the design does not allow for manual operations or easy reconfiguration of the payload. Next, we discuss other approaches to the ASV design and how their contributions have influenced our design. 

\invis{there are different end\hyp state goals driving their design, \textcolor{red}{N: and yet remains contributions that we have leveraged in our design.}}
Among the earlier designs of a small scale ASV was MARE by Girdhar \etal \cite{GirdharIROS2011}. Driven by low cost considerations, it enabled collection of visual data over shallow coral reefs and operated as a communication point in multi\hyp robot operations~\cite{Rekleitis2012b}. The design was based on the catamaran style with two electric motors that where controlled in a differential drive configuration. Battery powered, the range of operations was limited. Similar catamaran design with two electric motors have also the Kingfisher and the Heron Unmanned Surface Vehicle (USV) models from Clearpath Robotics\footnote{\url{https://www.clearpathrobotics.com/}} while portable, their range of operations is limited to one to two hours. 

In 2005, MIT's Curcio \etal  introduced their surface craft for oceanographic and undersea testing (SCOUT)\cite{curcio2005scout}. SCOUT is geared for oceanographic deployment based on an obstacle avoidance system working in conjunction with a remote palm device for high\hyp level mission control.  Their pioneering design and build of a truly unmanned boat set the stage for a variety of expansions of their original design. For our purposes, the electric drivetrain results in an increased draft to allow clearance for the electric motor head and propeller to displace water below the stern of the kayak.  As well, the mission specific sensor design and implementation offer us insight for our design to remain flexible and modular to accommodate larger and heavier instrumentation. Examples include sidescan sonar sensors and acoustic Doppler current profilers (ADCP) that are becoming prolific as the devices become more affordable. The small size limiting manual operation capabilities, low operating maximum speed of 5.6 kilometers per hour, and medium operation time of eight hours between charging are shortcomings of SCOUT that we seek to improve.

In 2008, Santa Clara University introduced their small waterplane twin hull (SWATH)\cite{mahacek2008development} ASV. SWATH employed two under the surface catamarans running  electric motors mainly for shallow and inland water operation. They originally deployed multibeam sonar as a tool for bathymetric mapping. Again in 2013, a Master's Thesis\cite{sclara} sought to improve the path following capabilities of SWATH in environments with wind and current present. While the results are successful in moderate conditions, its speed and deployment duration do not fit our needs for long distance deployment. However, their off-board control system inspires our design and implementation for customized control sequences for future mission specific tasks, such as object placement and retrieval.   

In December 2012, Rodriquez \etal wrote a comparison study\cite{rodriquez2012study} of existing ASVs for the specific purpose of measuring the environmental indicators that bear directly on climate change.  Throughout their study, they cover capabilities of satellites, weather balloons, RADAR, stationary buoy arrays, manned boats, autonomous underwater vehicles (AUV), and ASVs.  They compared each platforms capabilities of measuring wind speed, wind direction, water salinity, water temperature, barometric pressure, and oil mapping.  Their comparison of generic platform capabilities logically concluded that only manned boats and ASVs were capable of monitoring all indicators. Their report goes on to compare several AUV and ASV implementations with much insight gained from interviews with scientists and engineers from NOAA, WPI, Social and Environmental Research Institute (SERI), and the Applied Ocean Physics and Engineering Department from WHOI. Leveraging their conclusion that both manned boats and ASVs provide the greatest capabilities in their study led us to our requirement to keep our design flexible enough to support manned and unmanned operating modes.

In 2014, Fraga \etal introduced Squirtle\cite{fraga2014squirtle}, an autonomous electric catamaran for inland water environmental monitoring.  While their lack of passenger carrying capability, deeper draft due to propeller shafts and reliance on an electric source are not in line with our design goals, their methods for implementing a ROS node to provide autonomous control based on precise real time kinematic (RTK) GPS and inertial measurement unit (IMU) measurements, provide insight to our challenges with a maintaining a fully capable, self\hyp reliant platform.

Based on the payload, speed, and mission duration capabilities in the reviewed literature above, we decided that the WHOI approach was the best starting point for our development. Their ingenuity and pioneering approach to expand the capabilities of a commercial platform are what led us to select MOKAI as the base platform from which to build the ASV. From there, we seek to add modularity and flexibility to their design in order to provide a multipurpose platform.

\subsection{Design Goals\invis{N: [Maybe too detailed, can be summarized with only first two sentences?]}}
Our design differs from WHOI with respect to expanding capabilities to include long term deployments for inland waterways with highly dynamic currents and landscape. Our design considerations include physical platform modularity, sensor mounting versatility, and controls integration flexibility. To enable these end\hyp state goals, we researched and planned for robust communication and micro\hyp controller platforms modified and configured to operate in marine environments with safety features such as the ability to remotely kill \invis{and start}the engine. The physical layout must support on\hyp board manual operation to support environmental scientists requiring a level of supervision during data collection. The platform must include the capability to host numerous above water sensors such as cameras, LIDAR, anemometer, GPS, radar, and communication components. Also, the platform must be capable of hosting at least four underwater sensors such as depth sounders, bathymetric imaging transducers, water current sensors, and cameras. On\hyp board layout requirements include power planning for a 24 volt power source and plug\hyp n\hyp play distribution panel for 12 and 5 volt devices. In addition to the factory joystick controls, the on\hyp board footprint must include space for our autonomous and teleoperation control box, programmable control boards (PCB) servicing desired sensors and additional minicomputers. Space for on\hyp board companion minicomputers must be retained for our on\hyp line autonomous control interface using the Robot Operating System (ROS)\cite{quigley2009ros} as a framework for software development and standardized data collection. The specific components, placement, and integration of these components is covered in Section \ref{sec:usc_build} and illustrated in Figure \ref{fig:usc_jetyak}. 

\section{Autonomous Field Robotics Lab's Jetyak} \label{sec:usc_build}
\begin{figure}[t]
\centering
  \includegraphics[width=1\columnwidth]{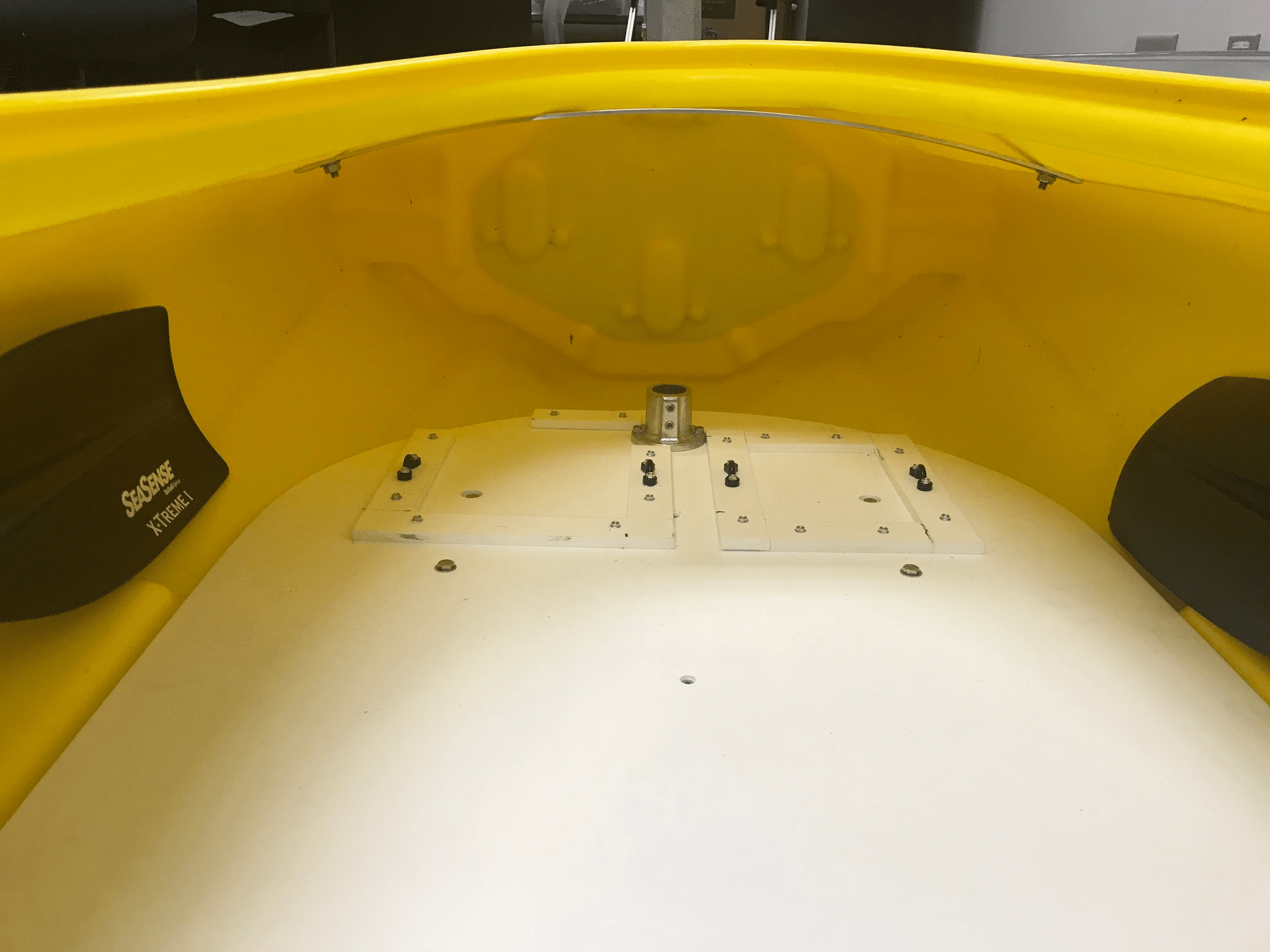}
  \caption{Front portion of marine grade starboard sub\hyp floor with footings for battery, control box, as well as topside and anchor mast ports installed.}
  \label{fig:hull}
\end{figure}
\subsection{Stock Platform}
Our base platform consists of the commercial  Mokai Es\hyp Kape\cite{mokai} boat, whose predecessor has been previously modified and termed the WHOI Jetyak by Kimball \etal \cite{whoiMokai2014}. The latest model ES\hyp Kape is 3.6 meters long and is propelled by a seven horsepower, four stroke, internal combustion engine costing \$5,400.  With its 9.8\hyp liter fuel reservoir, the ES\hyp Kape can operate at lower speeds for 18 hours and top speed for four hours before refueling is required.  Top speed with an average payload of 90 kilograms is 21.7 kilometers per hour, and the maximum payload capacity is 163 kilograms. Additionally, the factory ES\hyp Kape includes an improved jet drive with a clutch allowing the impeller to be stopped without stopping the engine, a modification that the WHOI team had to implement that we did not. As noted in WHOI's work, Mokai released their ES\hyp Kape model in 2014, which includes pulse width modulated (PWM) servos for both throttle and steering controls. This electronic control upgrade allows us to forgo developing electromechanical controls. This provides direct access to the servos controlling the throttle lever and steering nozzle by piggy\hyp backing on factory joystick controls and wiring harness. As a result, teleoperation and way\hyp point autonomous navigation controls are able to be directly implemented. In turn, this enables a ROS based control interface as a gateway to our research in developing an adaptive control system for operating in highly dynamic environments.

The remainder of this section describes the physical modifications to the platform, power distribution panel as well as the robotic controller integration. While our latest design and build is shown here, it should be noted that this design includes several lessons learned throughout the first four iterations of the modified Jetyak.   
\subsection{Physical Platform Modifications}
\begin{figure}[hb]
		\centering
		\begin{tabular}{ll}
		\subfigure[\label{fig:boat3}] {\includegraphics[height=0.305\textheight]{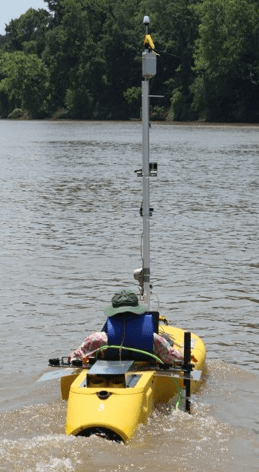}}&
			\subfigure[\label{fig:boat1}] {\includegraphics[height=0.305\textheight]{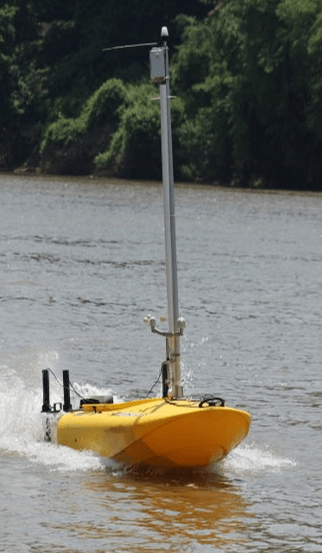}}
		\end{tabular}
			\caption{\subref{fig:boat3} Supervised Jetyak 3 with Ping DSP 3D sidescan imaging transducer, anchor light, communication hub, lidar, GPS, stereo camera, and monocular camera deployed. Engine wave guard and sensor splash guards are also installed and functioning.  \subref{fig:boat1}  Unsupervised Jetyak 1 with anchor light, communication hub, anemometer, 2 surface current and depth sonar sensors deployed.     
			\label{fig:man_unmanned}}
	\end{figure}
Our end\hyp state goals of maintaining manual operation capability and robust underwater and terrestrial sensor deployment flexibility translate to challenging spatial planning and layout considerations. Additionally, through lessons learned from initial deployments, three areas that require protection from the marine environment were identified.

\subsubsection{Interior Footprint}
When received, the inside hull of the Es\hyp Kape is a single layer of plastic maintaining the same shape as the exterior.  In order to mount boxes, plan for a mast, and keep wires off the bottom of the boat where water could collect, a sub\hyp floor of marine\hyp grade polymer starboard was constructed. This sub\hyp floor sits on the side steps of the hull and is fastened to the hull with stainless steel screws in the reinforced area of the bow, above the waterline. In addition, footings and tie downs were installed for the batteries and our electronics control box.  Finally, the base for the mast is added directly under the top mast port for added rigidity, illustrated in Figure \ref{fig:hull}. 

\subsubsection{Terrestrial Sensor and Communication Platform}
Another area that our design diverges from previous implementations is derived from our long\hyp range communication and robust terrestrial sensor requirements. In order to extend 2.4GHz and 915MHz radio communication range, we experimented with different materials and mast lengths until we found the optimal setup to be 2.4 meters of fiberglass pipe. This height allows for the Jetyak to be trailered on highways and while maximizing height for longer range line-of-sight capabilities. Its rigidity, low weight and electrically non\hyp interfering properties with the cables and antennas are desirable properties for this application.  Since this mast anchors to the bottom of the sub\hyp floor, we added a second lightweight PVC pipe to the outside to serve as a conduit for cables.  As seen in Figure \ref{fig:boat3}, the mast is capable of supporting the radio box, lidar sensor, GPS, stereo camera, and monocular camera. The flexibility of the mast mounting strategy is illustrated in Figure \ref{fig:boat1} where the Jetyak hosts an anemometer on the same mast.  
\subsubsection{Underwater Sensor Platform}
Again, our design goals were to develop a highly modular platform capable of deploying all types of sensors without the need to retrofit or make structural changes to the base boat. We decided to develop a strong, lightweight universal outboard mounting plate to permanently attach to the Jetyak. The complementary component to such a design is the vertical mounting poles that have a universal mounting ring welded to the bottom. The plate and pole design was delivered to a local water jet facility for cutting and welding the 6.35mm aluminum plate and brackets. As seen in Figure \ref{fig:man_unmanned}, each pole can be raised, lowered or removed independently according to researcher requirements. The underwater sensor in Figure \ref{fig:boat3} is the \SI{8}{\kg} 3DSS\hyp DX\hyp 450 sidescan transducer from Ping DSP\cite{ping}.
\subsubsection{Engine \& Electronics Water Protection} \label{sec:protect}
As we continued to develop and test the Jetyaks in rougher lake waters and faster moving currents, we learned quickly that protection for the air intake of the air cooled engine would be required. There are two ways that water can enter the engine compartment and air intake in our design.  First, and consistent with all Mokai stock platforms is the possibility of water from waves overflowing the top of the engine.  Using examples from our predecessors, we fabricated and installed a simple galvanized metal protection hood as visible in Figure \ref{fig:boat3} to guard against this hazard. The second hazard, due directly to our outboard sensor design, is from water deflecting up the sensor mounting poles into the air intake. This is overcome by cutting 3.175mm thick Lexan plastic to mount under the plate and extend forward and rearward of the plate to deflect water back away from the engine. Finally, as seen in previous implementations where humans are part of the payload, we integrated a windshield to abate spray from the front of the boat away from occupants and electronics.      
\begin{figure}[t]
\centering
\includegraphics[width=1\columnwidth]{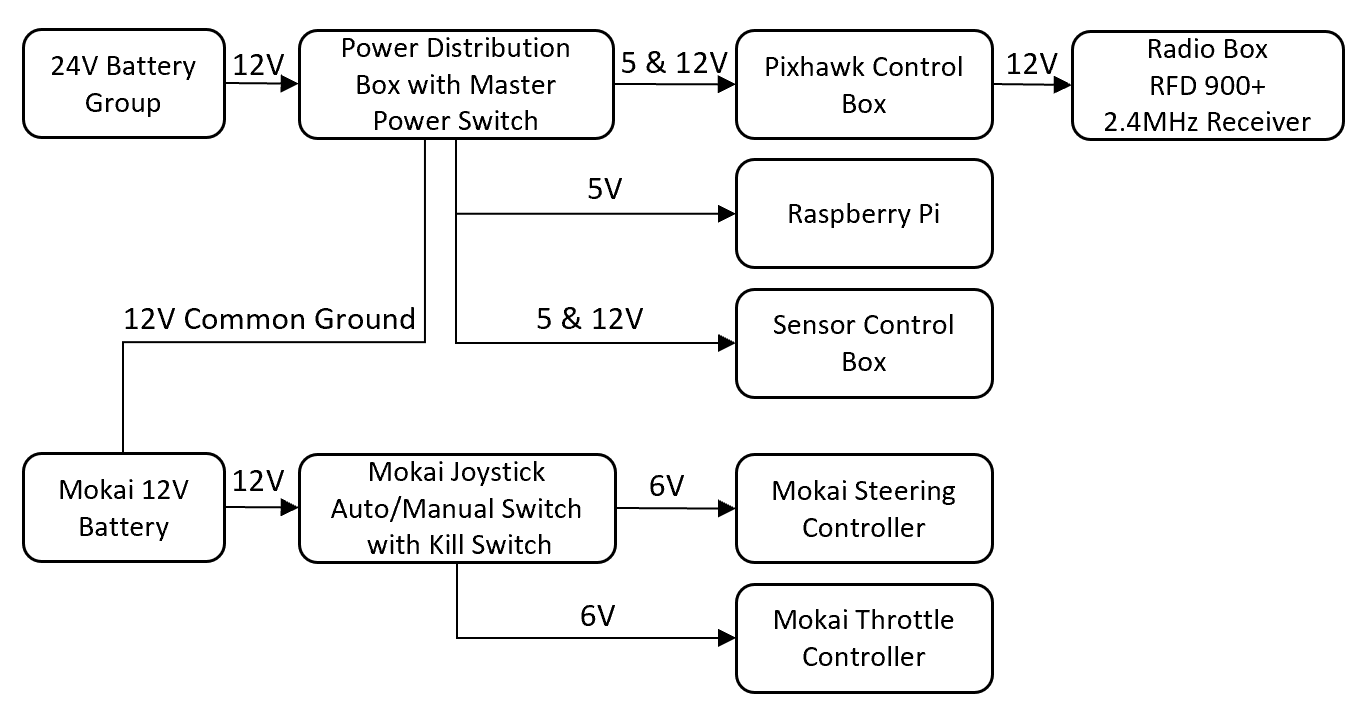}
  \caption{Power distribution diagram with our 24 volt power bank added to the Mokai factory system.}
  \label{fig:power_diagram}
\end{figure}
\subsection{Power Distribution}
To meet the requirements of many high end oceanographic sensors such as sonars, radars, ADCPs, and sidescan sonars, a 24 volt power source is required.  We accomplish this by connecting two 12 volt deep cycle batteries in series for direct wiring of 24 volt electronics. Figure \ref{fig:power_diagram} illustrates the power sources for each on\hyp board component. We provide 12 volts from one of the bank batteries to our power distribution panel.  Within the power distribution panel, we add an additional step down of the 12 volt line to five volts for our Pixhawk, Raspberry Pi, and Arduino PCB power supplies. This initiative is a result of a lesson learned after the addition of more than three sensors caused a confusing and cumbersome array of power and sensor wires, resulting in the loss of at least one Pixhawk PX4 and one Arduino UNO. A second lesson learned after witnessing some erratic servo behavior when controlling through the Pixhawk, was the requirement of ensuring a common ground ties both systems together. Since on a watercraft there is not a true ground but rather a floating ground reference, providing a common ground connection, ensures the factory servos and our added control system maintain the same zero voltage reference. Once the electrical bugs were identified and corrected, the integration of the power distribution panel resulted in a clean plug\hyp n\hyp play system which also provides better durability when deployed with a human on board.
\begin{figure}[b]
\centering
\includegraphics[width=1\columnwidth]{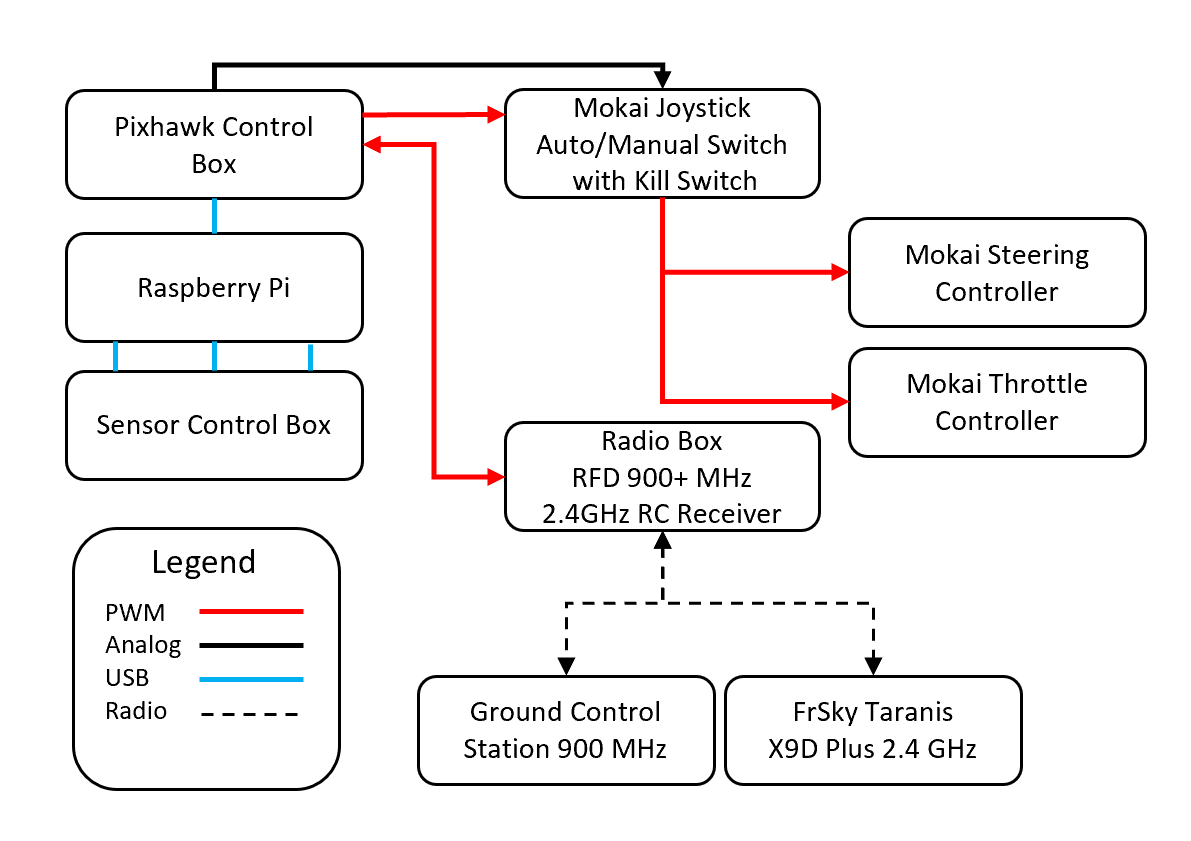}
  \caption{Jetyak controls architecture, illustrating the integration of factory, on\hyp board, and off\hyp board components.}
  \label{fig:controls_diagram}
\end{figure}
\subsection{Robotic Control Integration}
Our requirement to maintain five methods of operating the Jetyak required significant design and planning. The following subsections describe our design and build of the Jetyak with  a natural progression from preserving manual operation to remote control teleoperation and way\hyp navigation to our current work in developing adaptive controls on a ROS node. The schematics, PCB controller code, diagrams, pictures and configurations are included in our open source Jetyak tutorial page at the AFRL resource page \cite{doc}.

\subsubsection{Integrating with Factory Components}
In a worst\hyp case scenario, we ensure that the Mokai Jetyak maintains its factory manual operating capability. This drove our decision to place a manual/auto switch in the factory joystick control box that would always allow us to take over manual control of the boat. As illustrated in Figure \ref{fig:controls_diagram}, there are two intersections of our equipment with the Mokai's controls in the joystick box. 

The PWM wires connect the steering and throttle outputs of the Pixhawk to autonomous side of manual/auto switch and the factory PWM outputs of the joystick are attached to the manual side. At this point the factory 6 volt and ground connections are passed through with the manual/auto switch output on the factory harness to the servos. There are two advantages to this implementation. First, we reduce the footprint and exposure of our additional PWM carrying wires to the protected area in the factory joystick box. Second, by passing through the factory voltages, we eliminate the need to step up our five volt control voltage to the six volts required by the servos. 
\subsubsection{Remote Shutdown and Power Loss Safety Circuit}
Although our goal is toward a fully autonomous Jetyak, safe testing and deployment requires a method for remotely shutting down the Jetyak in case of emergency or eminent crash. \invis{And while we are at it, why not add the convenience of a remote start capability.} To accomplish this, we provide a parallel kill analog control connection to the factory circuits. Kill or shut down is accomplished through closing a loop which shorts the engine magneto to ground. We emulate this active low behavior through programming a digital channel on the Taranis radio and in the Pixhawk to normally operate in the high state, and when kill is activated, change to the low state. Using this output of the Pixhawk as the coil input for a relay results in the relay being energized during normal operation. When the Pixhawk signal goes low, normally closed contact is made in the relay. We provide this circuit in parallel to the factory system so that if either our system or the factory kill switch is activated the boat shuts down. It should be noted that this configuration also shuts the boat down when power is lost to the Pixhawk. As this remote safety feature may not be desirable when manually driving or recovering a Jetyak, we provide a physical override switch on the side of the Pixhawk box to effectively disable this feature.

\begin{figure}[t]
\centering
\includegraphics[width=1\columnwidth]{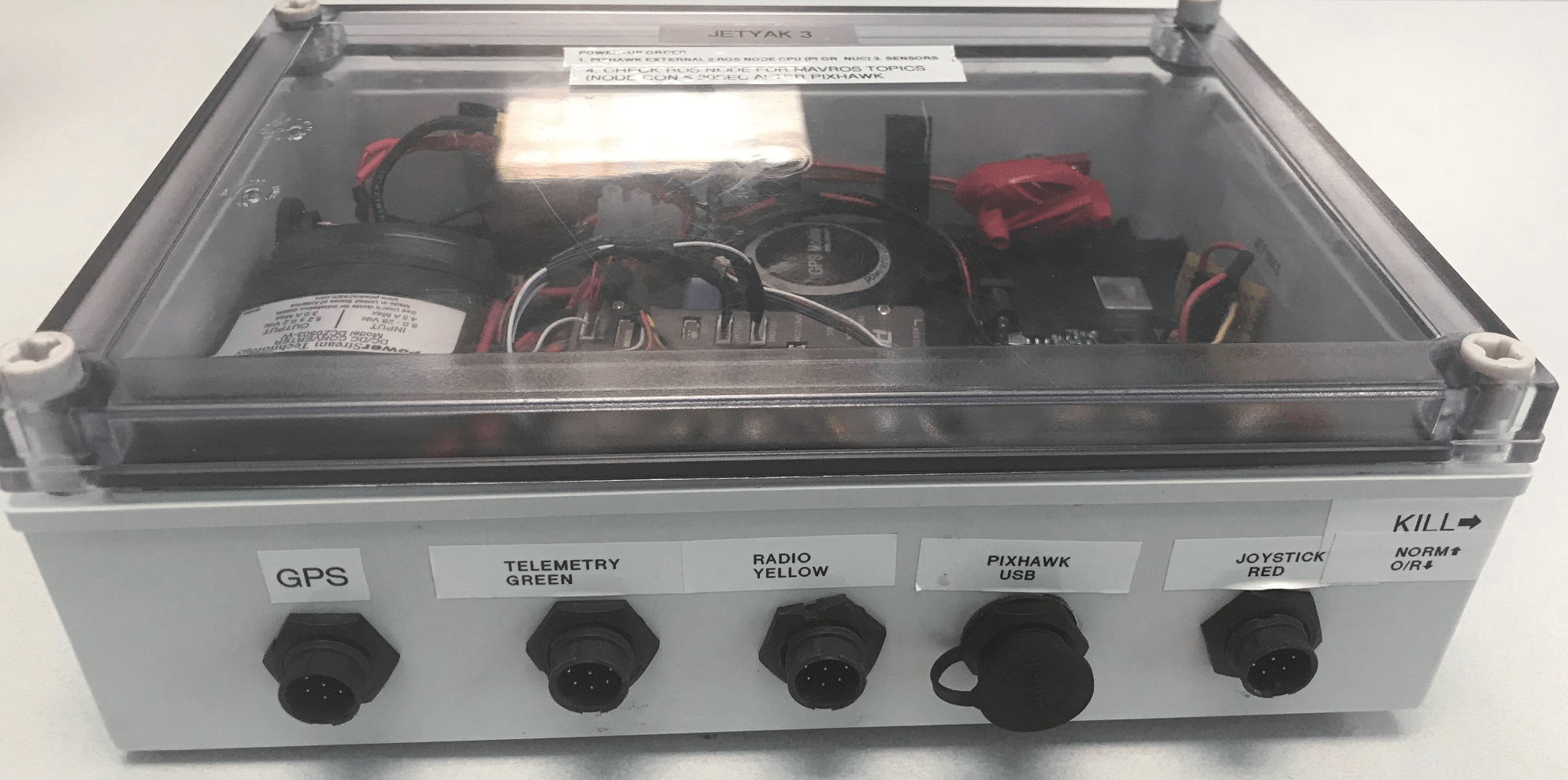}
  \caption{Pixhawk Box with power conditioner, Pixhawk PX4, Arduino Uno, GPS, joystick outputs, remote control interface, MAVLink \hyp  MAVROS node proxy inputs/outputs collocated.}
  \label{fig:pixhawk}
\end{figure}
\subsubsection{Baseline Teleoperation and way\hyp point Navigation}
At the heart of our design, we selected 3DR Pixhawk 1 running PX4 on the NuttX\cite{nuttx} operating system, along with the ArduPilot Software Suite\cite{ardupilot} to enable teleoperation and way\hyp point navigation capabilities, shown in Figure \ref{fig:pixhawk} in its Pixhawk box. This version of the Pixhawk includes an internal compass and external I2C (inter\hyp integrated circuit) compass port for an external compass, included with most GPS antennas. Configuration of the PixHawk as a Rover \cite{Rover} allowed us to start from a point where the throttle and steering servos on the stock Mokai Jetyak are directly imitable. Both the Hitec HS\hyp 5485HB\footnote{\url{https://hitecrcd.com/products/servos/discontinued-servos-servo-accessories/hs-5485hb-standard-karbonite-digital-servo/product}} throttle linkage servo and the Torxis i04903\footnote{\url{https://gearwurx.com/product/torxis-industrial-outdoor/}} steering servo are controlled by modifying the Pulse Width Modulation (PWM) values in the PixHawk to match their operating specifications.  The latter steering servo is internally controlled by the Polulu Jrk21V3 USB motor controller, which allows the user to access the PWM cutoffs and allows direct calibration between the mechanical steering angle and the input signal. This results in one valuable lesson learned, in that electromechanical devices with identical specifications, operate differently given the specification tolerances.  In order to overcome these challenges, it is extremely beneficial to measure the factory PWM output widths at minimum, maximum and center with an oscilloscope prior to attempting to teleoperate the servos.  This allows the parameters for the throttle and steering output channels on the Pixhawk to be properly aligned from the start. Another lesson learned is to ensure the orientation and calibration of the compass(es) are accurately completed, since different manufacturers of external compasses may assume a non-forward mounting orientation. 

The second enabling component to our implementation is the Taranis X9D plus radio system, which offers great flexibility in programming explicit controls of up to 16 channels when paired with the Taranis D8R\hyp XP receiver. In our case, we use the community standard of channel one to control steering and channel three to control throttle. We use channel six to provide our teleoperated kill capability, and channel five to control the mode of operation. The last step to enabling remote control operation of the Jetyak is to configure the appropriate channel outputs in the Pixhawk to match behavior characteristics required for the controlled device, \eg servo PWM minimum and maximum thresholds, PWM trim (neutral) position and forward/reverse direction.  The highly modular programming interface allows for fully customized servo and switching controls based on logical functions, making the remote control capabilities very granular. For instance, there are five desired control modes for our application from manual progressing to guided or off\hyp board control which cannot be programmed with a single 3\hyp position switch. The X9D is capable of assigning a distinct PWM signal reflecting the logical result of the positions of a 3\hyp position and 2\hyp position switch. This example is completed when the Pixhawk is programmed with the corresponding mode functions for the received PWM signal on that channel. As a result, the Jetyak is capable of being supervised when testing autonomous capabilities and can always be overridden, which is a necessary feature when conducting field trials in the public domain. Next, we will describe the platform enhancements we have added to enable greater autonomy beyond way\hyp point following.
\subsubsection{ROS Integration}
Initially, the platform was used to conduct preplanned missions, collect data, and return to its home location. To accomplish this task, we only need a common time source, location and the desired sensor measurements to be synchronized. Since the goal is to produce an autonomous Jetyak for deployment in highly dynamic environments, we preplanned implementation of a system that could collect the information in a format that would be available for online usage. Luckily, there exists a well\hyp supported, open\hyp source solution readily available to support our needs, ROS. 

ROS is a robust middleware providing a framework for publishing and subscribing to topics and messages between different process, low\hyp level device controllers and on\hyp board computers. In addition, it provides a package management environment enabling add\hyp on packages such as MAVROS to interface with many off\hyp the\hyp shelf controllers such as the Pixhawk. This allows access and integration with IMU, GPS, heading, velocity, pose and several other Pixhawk telemetry topics. These topics are then published by the on\hyp board ROS node through USB connection on its host Raspberry Pi or Intel NUC. ROS also accommodates the addition of our depth sonar, current speed, and anemometer measurements directly into the same ROS framework. We have included depth, wind, and current sensors as a standard component to our Jetyak design, enabling operation in highly dynamic environments. 

\invis{[should we expand on what are those methods instead of general information?]}Lastly, in order for the Jetyak to use sensor measurements for on\hyp line path planning, the ROS framework provides an integration of sensing and acting commands. \invis{we require a method for executing the decisions and movements resulting from our environmental interpretation.} More specifically, ROS integration provides a topic publishing conduit for sending general navigation as well as channel\hyp level steering and throttle control commands directly to Pixhawk using MAVROS and the MAVLink protocol. The specific approach is up to the specific application.  

\subsubsection{Robust Communication}
AFRL's Jetyak maintains three forms of communication to allow interfacing and programming at different levels and distances.  Short\hyp range communication is maintained through 802.11g wireless ad\hyp hoc connectivity to the NUC and Pi devices. Remote teleoperation and low\hyp level telemetry communication is provided through the FrSky Taranis X9D Plus transmitter to D8R\hyp XP receiver radio link operating in 2.4MHz spectrum.  While the long\hyp range communications is provided through RFD 900+ MHz modems, with one as a base station and one modem per deployed Jetyak node. With the addition of our 2.4 meter mast, we have been able to extend our line of sight communication with the base station to 2.8 kilometers. Note: While not a best practice, it is possible to allow the Pixhawk to continue its programed mission without this communication link.  
\subsubsection{Initial Tuning Requirements}
Initial deployment and testing included manual refinement of steering and throttle servo proportional, integral, and derivative (PID) controls to establish the reliable path following capability of the ASV. The starting point and manual procedure for this tuning is included on the AFRL's Jetyak tutorial\cite{doc}.  
\invis{A good starting point for steering PID parameters are shown below:
\begin{itemize}
\item STEER2SRV\_D 0.005
\item STEER2SRV\_I 0.2
\item STEER2SRV\_P 2
\item WP\_RADIUS 5 meters
\end{itemize}
The manual procedure we used to tune the PixHawk PID coefficients in calm water  follows: 
\begin{enumerate}
\item ASV turns too slowly, increment proportional gain.
\item ASV oscillates more than three times before finding target line, decrement proportional gain.
\item ASV oscillates more than 0 and less than 3 times before finding target line, increment derivative gain.
\item ASV oscillates small amounts at high frequency, decrement derivative gain.
\item ASV still oscillates at lower frequency, decrement integral gain. 
\item ASV starts to turn before reaching way\hyp point, decrement way\hyp point radius.
\item ASV starts to turn too far after reaching way\hyp point, increment way\hyp point radius.
\end{enumerate}
Note: We began by adjusting the proportional coefficient} 
Once tuning the PID controller coefficients is complete, we were able to deploy a single Jetyak on way\hyp point tracking missions. These missions were created from a Dubins vehicle grid search coverage algorithm developed in UofSC's AFRL\cite{icra2018} shown in Figures \ref{fig:waypoint} and \ref{fig:single_trajectory}.

\section{Lessons Learned}\label{sec:lessons}
\begin{figure}
\centering
\includegraphics[width=1\columnwidth]{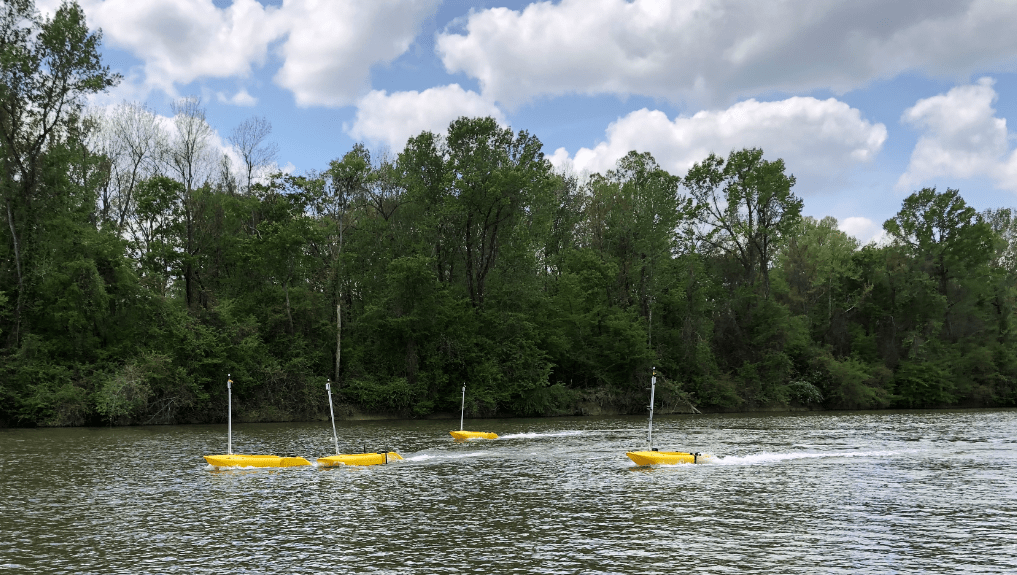}
\caption{\label{fig:4-jetyak}Four Jetyaks operating autonomously in the Congaree River near Columbia, SC. }
\end{figure}
In addition to the implied lessons learned in Section \ref{sec:usc_build}, we will discuss other valuable lessons learned during the building and deployment of a fleet  of Jetyaks. First, the lessons we have learned in building the first five Jetyaks of UofSC's fleet will help anyone seeking to develop their own Jetyak avoid some of the pitfalls that cost us time and money in terms of replacement costs and repair time. The second general area, often overlooked, is the lessons learned during field testing any platform in the real world. In our specific design and implementation, we learned some valuable lessons associated with the outboard sensor mount that must be understood and overcome to collect reliable, consistent data. Lastly, time and resources should be allotted for maintaining the fleet. Ignoring these lessons often costs precious time, especially when considering the logistics involved with hauling and launching one or more boats.
\subsection{Building a Fleet instead of a Single Vessel}
\subsubsection{Electromagnetic Interference} When working with an internal combustion engine, the magneto introduces interference. In the first iteration of building a Jetyak, we implemented a separate auto/manual switching box that used the factory joystick outputs and our Pixhawk control outputs as its inputs and used our in\hyp house fabricated cables to connect to factory ports on the engine bay. After the second day of testing continued to produce unpredictable behaviors, we began monitoring the switch box outputs with an oscilloscope to find that when we switched the system to auto (Pixhawk) signals, an inordinate amount of noise was introduced. This can have catastrophic effects when working with PWM signals. If the last signal sent happens to correspond to the servo manipulation for accelerate, then the servo will continue to hold that position until overridden. In order to rectify this, we designed a system with the same signals, but this time, eliminating any non\hyp factory wiring beyond the outputs of the control box. The results were much cleaner signals in both operating modes and stable behavior.
\invis{\begin{figure}[b]
\centering
\includegraphics[width=1\columnwidth]{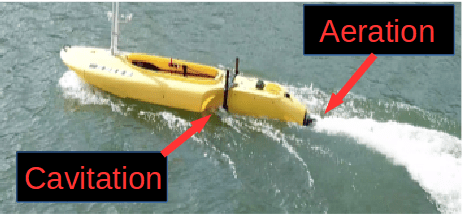}
\caption{\label{fig:cavitation}The effects of cavitation from the hull and aeration from the sensor and mount acting on a single paddle wheel speed sensor on Jetyak 1.}
\end{figure} }
\begin{figure*}[t]
		\leavevmode
		\begin{tabular}{ccc}
		\subfigure[\label{fig:waypoint}] {\includegraphics[height=0.23\textwidth]{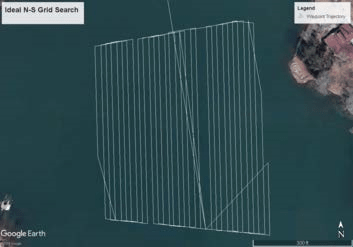}\label{fig:s1}}&
			\subfigure[\label{fig:single_trajectory}] {\includegraphics[height=0.23\textwidth]{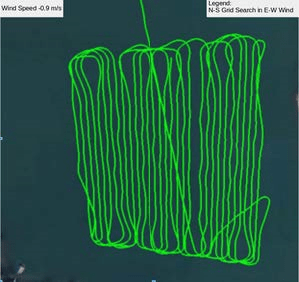}}&
			\subfigure[\label{fig:multiple_trajectory}]{\includegraphics[height=0.23\textwidth]{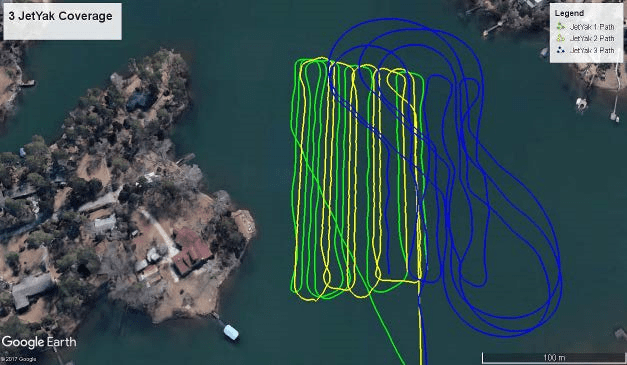}}
		\end{tabular}
			\caption{(a) An ideal North\hyp South grid search mission used to provide a baseline for measuring tracking performance with minimal turning radius of 5 meters.  (b) Trajectory for grid search conducted by a single Jetyak with tuned PID coefficients. (c) Trajectories for three Jetyaks searching their respective areas of responsibility according to mission resulting from implementation of our multi\hyp robot coverage algorithm for Dubins vehicles.   
			\label{fig:search_behavior}}
	\end{figure*}
\subsubsection{Maintaining Compass Accuracy} The Pixhawk is capable of maintaining two compass headings, its internal compass and an external. Depending on the quality of the external I2C compass purchased, in our case 3DR, it is often beneficial to assign priority to externals. In order to compare the reliability of two compasses, simply select each compass as the primary compass in Mission Planner and select the one that drifts the least in a stationary environment. Finally, when the compass is mounted in an orientation not aligned with the Jetyak, it is essential to input the axes orientation and calibrate the compasses in Mission Planner to ensure the proper offsets are maintained.
\subsubsection{Repeatability \& Quality Control} A large scale project such as this will thoroughly test any lab's methods for ensuring efficiency and best practices. Recognizing differing methods for maintaining best practices, some examples of areas in the project that will consume time and money follow:  
\begin{itemize}
\item {\bf Standardized Wiring Color Scheme:} Develop a standard wiring color scheme beginning with the factory joystick box scheme. 
\item {\bf Documentation Standards \& Sharing:} Developing a standard for real-time collaboration and sharing of design changes is crucial when lab turnover occurs.
\item {\bf Adopting Industry-like Quality Assurance \& Control Standards:} Establishing a hierarchy within the build team will save several hours of troubleshooting the dozens of circuits required to make this architecture function. 
\end{itemize}

\subsection{Real\hyp world Deployment} 
Field deployment of the Jetyak brought a new level of learned lessons to our team, especially in the domain of highly disrupted marine deployments. These lessons fall into three categories expanded below: field trial deployments, outboard sensor deployment, and maintenance. 

\subsubsection{Field Trials} The logistics required to plan and safely execute field trials with one or multiple Jetyaks cannot be understated. After our first attempt to deploy a Jetyak at Lake Murray, SC in a generally calm and stable environment, we developed a comprehensive startup checklist to ensure all components were operational in a sequential manner. The general component groups we test are the steering, throttle and kill operations as well as ground control station connectivity in manual, remote control, and autonomous modes. Each aforementioned test is carried out with the boat engine off as well as running for thoroughness. Before deeming the Jetyak ready for launch, we ensure our ROS node is operational and receiving all required MAVROS and sensor topics. Our field trial log sheet is included as an appendix to AFRL Jetyak tutorial\cite{doc}. \invis{The first section of checklist is devoted to ensure proper maintenance is completed. Lessons learned concerning maintenance are covered below in Subsection \ref{sec:maintain}. Then employing the crawl, walk, and run principle, we test each component with the Jetyak on the trailer, engine off, and switches on. This crawl step ensures the protection of factory components if something is connected or installed incorrectly. Then we test the same components after starting the boat in manual drive mode. At this point, we also ensure that our ROS node is operational and receiving all required MAVROS and sensor topics. At this point the boat can be shut down and is prepared for launch. Our field trial log sheet is included as an appendix to AFRL Jetyak tutorial \cite{doc}.}
\subsubsection{Outboard Sensor Orientation}
Other physical phenomena we contend with are the resulting cavitation and aeration effects of moving a body through water. Cavitation must be considered when deploying physical measurement sensors such as current sensors, and aeration will quickly become the enemy of sonar based sensors, causing erratically high or undefined readings. \invis{Effects of aeration and cavitation are illustrated in Figure \ref{fig:cavitation}. } In our case, several trials were required to find the best location and orientation with relation to the ASV to ensure accurate readings. Generally, the sensor needs to be mounted slightly deeper than any hull of the boat traveling in\hyp front of the sensor, and away from the disruption area of the propulsion system. In addition the mounting pole of the sensor should be mounted behind the sensor. These two tactics allow unperturbed water to cover the bottom of the sensor. Planning for and reducing sensor exposure to the effects of cavitation and aeration will save much frustration and time lost in trips to and from the launch site for future builders.

\subsubsection{Maintenance} \label{sec:maintain}
Proper routine maintenance of the Jetyak will ensure proper mechanical operation for the next deployment. Tasks such as topping off fuel, checking and changing the oil when required, charging batteries, greasing the drive shaft coupler can be completed days or weeks prior to the next deployment. These tasks are also captured on our startup checklist included in the tutorial. 

Lastly, if excessive water does make its way into the engine air intake, impromptu maintenance must take place otherwise catastrophic failure may occur. When this happens, the engine should be stopped and the boat returned to safety where the engine can be removed and the oil changed several times until the milky appearance has disappeared. Due to the modular design of Mokai's lock and pin assembly, it is feasible to include an extra engine or engine box as part of the field trial support package to reduce downtime if this does occur.

\section{Results}\label{sec:results}
In this section, we provide some examples of field deployments we have completed with the Jetyak. Initial sensor payload includes different combinations of Ping DSP sidescan sonar, Humminbird Helix 7 sidescan, Velodyne lidar, stereo camera, depth sonar, anemometer, and current sensors. The measurements and predictive data illustrated here is still under analysis and development for future improvement goals.  None\hyp the\hyp less, they give some intuition into the utility and versatility of the Jetyak for exploration and task focused data collection. Real world data collection results include sonar, anemometer, and surface current measurements in stable environments on Lake Murray, SC and the highly dynamic environment of the Congaree River near Columbia, SC. In addition to the raw measurements, initial mapping and prediction capabilities are illustrated for close temporal planning.
\subsection{Stable Environment Deployments}
\begin{figure}[t]
\centering
\includegraphics[width=1\columnwidth]{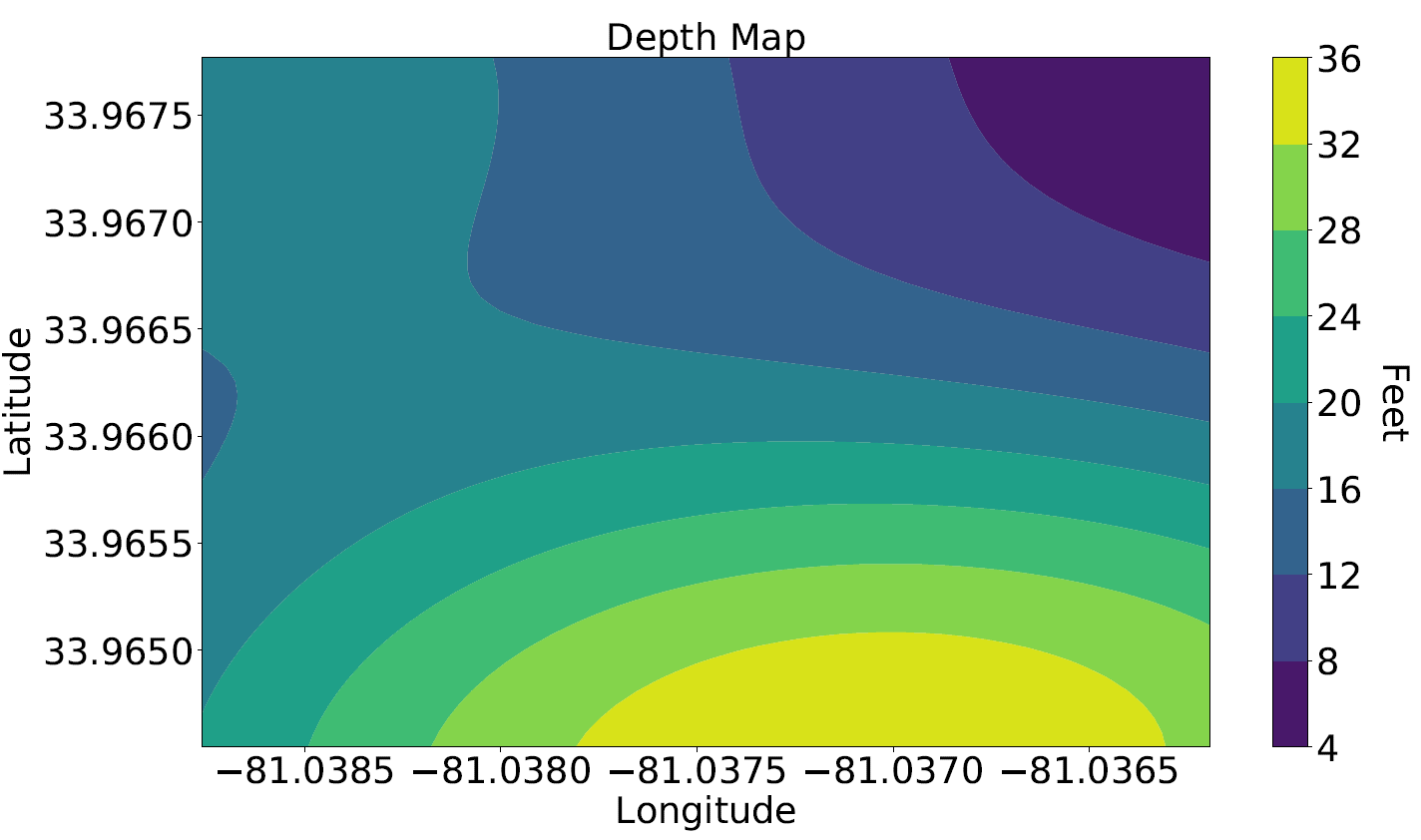}
\caption{\label{fig:depth}Depth map of \SI{1}{\km} portion of Congaree River, SC.}
\end{figure}
\begin{figure}[b]
\centering
\includegraphics[width=1\columnwidth]{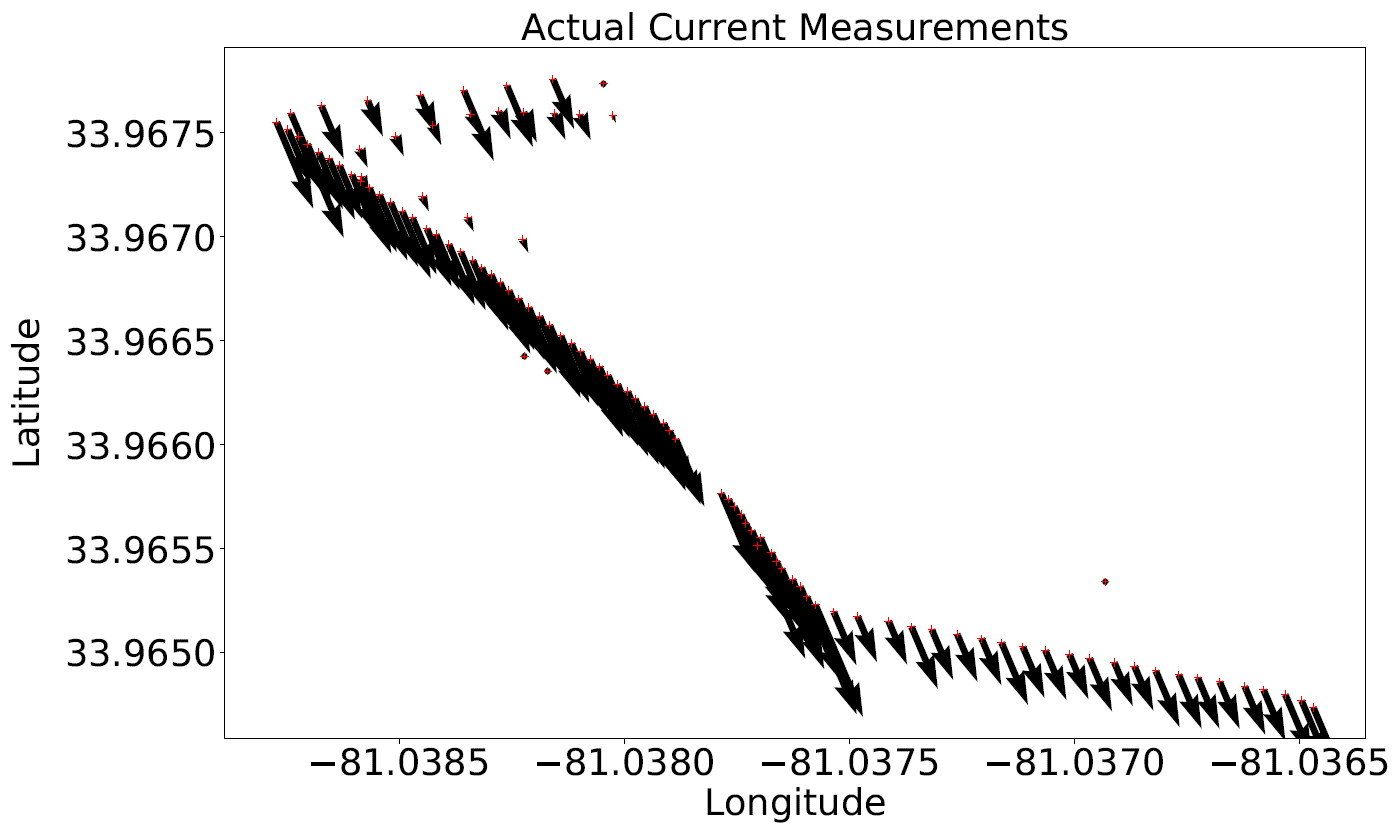}
\caption{\label{fig:current}Actual current direction and intensities after transformation from boat reference frame to world reference frame taken on Congaree River during flood stage currents.}
\end{figure}
\begin{figure}[t]
\centering
\includegraphics[width=1\columnwidth]{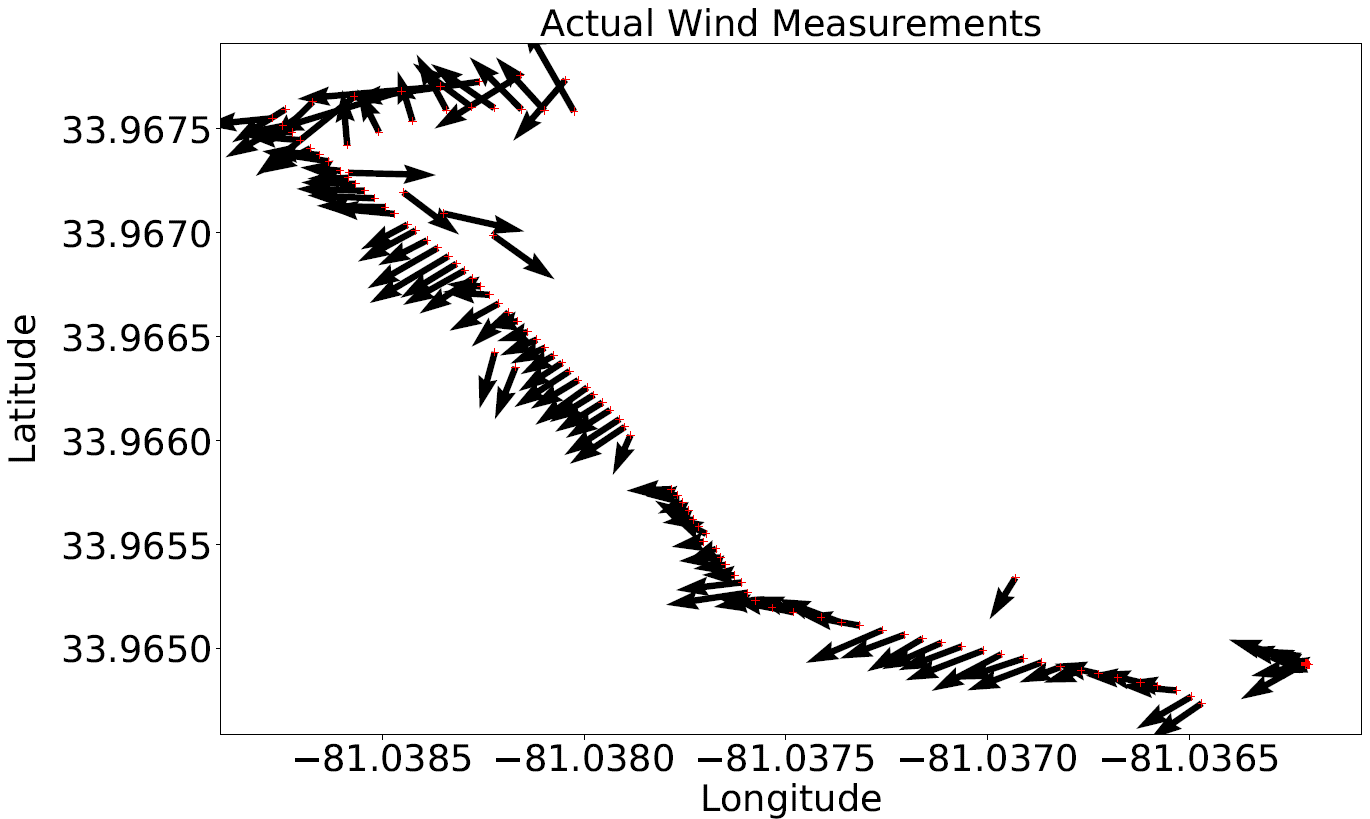}
\caption{\label{fig:wind}Actual wind direction and intensities on Congaree River after transformation from the boat reference frame to the world reference frame.}
\end{figure}
Initially, we deployed a single Jetyak on Lake Murray in South Carolina with a single sonar depth sensor to test the Jetyak's performance from launch to autonomous operation to data collection and logging. Afterwards, multiple Jetyaks were used as an experimental setup in the work by Karapetyan \etal\cite{icra2018} for deploying a Multi\hyp Robot coverage algorithm with Dubins kinematic constraints. The trajectories of each robot are illustrated in Figure \ref{fig:search_behavior}. The coverage was performed locally, by each Jetyak tracking preassigned way\hyp points programmed in the mission planner. Given the recent addition of Jetyak 3 to the fleet at the time of this experiment, the lack of time to tune the PID coefficients resulted in Jetyak 3's erratic behavior in Figure \ref{fig:multiple_trajectory}. The resources required to maintain accurate tuning and overcome this behavior for the entire fleet along with how naturally occurring disturbances (wind, current) adversely affect our ASV platform has motivated our future work in adaptive controls. 
\subsection{Highly Dynamic Environment Deployments}
As we extended our platform to operate and collect data in more volatile environments, we are able to complete our data collection goals in water currents reaching nearly \SI[per-mode=symbol]{3}{\m\per\s}. The PID controller is able to track way\hyp points against the current by slowly working against the current until it reaches the desired point. However, the Jetyak misses several way\hyp points that are downstream or cross stream where faster moving surface currents exist. This experimental realization has reinforced our desire for future work in adaptive controls. Figures \ref{fig:depth} through \ref{fig:wind} illustrate the depth, current and wind measuring capability of the Jetyak in these adverse conditions. 

In addition, we have successfully deployed sidescan imaging sensors similar to the one on the MIT SCOUT using the modular poles without modification to our universal mounting bracket.
\section{Future Work}\label{sec:future_work}
Our future work builds on the demonstrated capability to collect measurements of depth, wind, current, side scan images, lidar data, and stereo camera images, by enabling on\hyp line methods for control as an augmentation to the PID controller. Yang \etal \cite{2014trajectory} published similar work focused on reactive controls once the phenomenon has affected the ASV's course. Expanding on their work with ocean\hyp going vessels, we focus on deployments in ports, tributaries, canals and rivers to enable exploration and monitoring of remote waters. In addition to providing the ability to generate models of the environment, we are exploring Gaussian Process based techniques to predict and model current and wind disturbances in short temporal windows to enable proactive controls for deployments in highly volatile situations. The long term goal is to provide a comprehensive set of hardware and software that allows scientists to easily field such vehicles for autonomous and robust deployments in a wide array of marine environments. 

\section{Conclusion} \label{sec:conclusion}
In this paper, we have shown the design and build of AFRL's Jetyak including the design considerations, components, build details, and lessons learned. The AFRL Jetyak is the result of customizing a commercially available Mokai Es\hyp Kape, including design considerations and comparisons to previous pioneers from WHOI, MIT, WPI, and Santa Clara University in their similar implementations. We demonstrate the utility of our design and build in demonstrations in both stable environments as well as highly dynamic environments. We illustrate our future work with this platform through identifying its current limitations of maintaining accurate trajectories in environments with high winds and surface currents; seeking to provide a solution that will allow deployment in such environments through adaptive controls. Finally, we provide a publicly available tutorial~\cite{doc} \invis{with component lists, vendors, costs, and pictures}to enable interested marine domain researchers to duplicate the presented system. 

\invis{Compared to other implementations, the outboard sensor mount platforms enable the UofSC Jetyak to remain flexible. This allows it to support both autonomous operation and human\hyp on\hyp board operation. In addition, the modular design of terrestrial and underwater sensor mounts allows the ultimate customization for a wide array of applications. Again, compared to electric implementations, the gasoline engine and improved factory design of the Es\hyp Kape provide the ability to deploy large payloads for longer distances and durations than previous implementations.    
Finally, we have illustrated our future work with this platform through identifying its current limitations of maintaining accurate trajectories in environments with high winds and surface current speeds. Through our continued research into short\hyp term dynamic monitoring, we seek to provide a solution that will allow deployment in such environments through adaptive controls.}

\section*{Acknowledgement}
This work was made possible through the generous support of National Science Foundation grants (NSF 1513203, 1637876). The authors would also like to thank Scott White for providing the Ping DSP sidescan sonar for payload testing. 
\bibliographystyle{IEEEtran}
\bibliography{IEEEabrv,refs}

\end{document}